\documentclass[a4paper, 10pt, conference]{ieeeconf}
\usepackage{cite}
\usepackage{amsmath,amssymb,amsfonts}
\usepackage{graphicx}
\usepackage{textcomp}
\usepackage{xcolor}
\usepackage[noabbrev]{cleveref}

\usepackage{algorithm}
\usepackage[noend]{algpseudocode}

\usepackage{caption}
\usepackage{subcaption}

\usepackage{printlen}

\usepackage{todonotes}

\usepackage[acronym, shortcuts]{glossaries}

\usepackage{booktabs}

\usepackage{titlesec}

\definecolor{orange}{HTML}{e69f00}
\definecolor{blue}{HTML}{0072b2}

\newacronym{drl}{DRL}{deep reinforcement learning}
\newacronym{pacmap}{PaCMAP}{Pairwise Controlled Manifold Approximation Projection}
\newacronym{rl}{RL}{reinforcement learning}
\newacronym{sac}{SAC}{Soft Actor-Critic}
\newacronym{ddpg}{DDPG}{Deep Deterministic Policy Gradient}
\newacronym{pca}{PCA}{Principal component analysis}
\newacronym{gan}{GAN}{Generative Adversarial Network}
\newacronym{relu}{ReLU}{Rectified Linear Unit}

\begin{document}

\title{Deep Reinforcement Learning Behavioral Mode Switching Using Optimal Control Based on a Latent Space Objective}

\author{Sindre Benjamin Remman, Bjørn Andreas Kristiansen and Anastasios M. Lekkas
\thanks{The authors are with the Department of Engineering Cybernetics, Norwegian University of Science and Technology (NTNU), Trondheim, Norway. {\tt\small \{sindre.b.remman, bjorn.a.kristiansen, anastasios.lekkas\}@ntnu.no}}
}

\maketitle

\begin{abstract}


In this work, we use optimal control to change the behavior of a deep reinforcement learning policy by optimizing directly in the policy's latent space. We hypothesize that distinct behavioral patterns, termed behavioral modes, can be identified within certain regions of a deep reinforcement learning policy's latent space, meaning that specific actions or strategies are preferred within these regions. We identify these behavioral modes using latent space dimension-reduction with \ac*{pacmap}. Using the actions generated by the optimal control procedure, we move the system from one behavioral mode to another. We subsequently utilize these actions as a filter for interpreting the neural network policy. The results show that this approach can impose desired behavioral modes in the policy, demonstrated by showing how a failed episode can be made successful and vice versa using the lunar lander reinforcement learning environment.

\end{abstract}


\section{Introduction}

\Ac*{drl} has emerged as a powerful paradigm for solving complex decision-making problems, ranging from playing video games \cite{badia2020agent57} to autonomous vehicle control \cite{kiran2021deep}. At the heart of \Ac*{drl}'s success is the ability of neural network policies to learn and adapt to diverse environments through interaction and feedback. However, understanding and interpreting the behavior of these policies remain challenging due to their inherent complexity and the opaque nature of neural networks.

Recent papers such as \cite{salzmann2023real} have shown how neural networks can be used for optimal control, using them as a way to approximate dynamical models and subsequently use them as part of a model-predictive control framework.

This paper analyzes the latent space of neural network policies using the dimension-reduction method \Ac*{pacmap} \cite{wang2021understanding}. In this case, latent space refers to the activations after the penultimate layer in the neural network policy, visualized in \Cref{fig:latent_space_network}. We observe that the policies we consider perform distinct behaviors in separate parts of the latent space. More formally, these neural network policies often perform a specific behavior within certain locations, or hypervolumes, in the policy's latent space. We call these hypervolumes and their corresponding behavior for \emph{behavioral modes}. We identify behavioral modes in neural network policies trained using \ac*{drl} and subsequently move the policies between behavioral modes using optimal control with a latent space objective function. More specifically, the contributions of this paper are:

\begin{itemize}
    \item We use two-dimensional embeddings of \ac*{drl} policies' high-dimensional latent space generated using \ac*{pacmap} to find situations in different episodes that are initially close in latent space but have different outcomes.
    \item We employ optimal control to find an action sequence that moves the environment's state latent space projection closer to a desired part of the latent space, where we hypothesize a desired behavioral mode exists.
    \item We show results indicating that by moving the environment to a state whose latent space representation is closer to the desired latent space location, we have effectively changed the agent's behavioral mode. 
\end{itemize}


\section{Preliminaries}\label{sec:preliminaries}

\subsection{PaCMAP}

\Ac*{pacmap} is a dimension reduction method designed to preserve the original data's global and local structure in the lower-dimensional embedding. The method was introduced in \cite{wang2021understanding}, where the authors compare and analyze several dimension-reduction methods. Based on the insights gained from this analysis, the authors discover helpful design principles for the loss functions used by successful dimension reduction methods. The authors subsequently use these design principles to design the \Ac*{pacmap} method. 

To enable the low-dimensional embedding to preserve the higher-dimensional data local and global structure, \Ac*{pacmap} uses three different types of point pairs: \emph{neighbor pairs}, which are points that are closest in the original data space, with distance defined using a scaled Euclidean distance function; \emph{mid-near pairs}, which are on average closer in the original data space than random points from the data set; and \emph{further pairs}, which are found by sampling non-neighbors. The loss function used in \Ac*{pacmap} is

\begin{align*}
    Loss^{PaCMAP} =\text{ }&w_{NB} \sum_{\text{$i$,$j$ are neighbor pairs}} 
    \frac{\tilde{d}_{ij}}{10+\tilde{d}_{ij}}\\ + &w_{MN} \sum_{\text{$i$,$j$ are mid-near pairs}} 
    \frac{\tilde{d}_{ij}}{10000+\tilde{d}_{ij}}\\ + &w_{FP} \sum_{\text{$i$,$j$ are FP pairs}}
    \frac{1}{1+\tilde{d}_{ij}},
\end{align*}
where $\tilde{d}_{ij}=||\mathbf{y}_i - \mathbf{y}_j||^2$, and $\mathbf{y}_i$ is the position of data point $i$ in the lower-dimensional embedding. $w_{NB}$, $w_{MN}$, and $w_{FP}$ are the weights for the three graph components. These weights are changed throughout the optimization process using a pre-defined strategy. The optimization process aims to preserve the original data's structure in the lower-dimensional embedding by minimizing the loss function, with the original data space's spatial relationships classified using the three different types of pairs described above. 

An important point to note is that even though \Ac*{pacmap}'s lower-dimensional embedding of the data can enable important insights into the data, the positions of the points themselves in the low-dimensional embedding are arbitrary; it is only the relationships between where the points are placed which is important. The dimensional axis in the low-dimensional embeddings does not have any units since the low-dimensional embedding combines the original dimensions in highly nonlinear and variable ways depending on the specific points and their neighbors. 

\subsection{Optimization in Spaces Defined by Neural Networks}

Solving optimal control problems in neural network spaces is a challenging task. An example of the issues faced is the nonlinearities introduced by the neural network activation functions, which potentially lead to discontinuities in the representation, depending on how much care is taken to make the problem fit for optimization. The natural high-dimensional nature of the problem and the non-convex features lead us to solve our problem as a nonlinear program (NLP), choosing the solver IPOPT due to its features on large-scale problems \cite{wachter2006implementation}. The nonlinearities from the activation functions could cause ill-posed problems, that is, problems where the solver cannot find a proper minimum. During our investigation, we discovered that the solver could not reach solutions while using the exact Hessian. When changing to the quasi-Newton method available to us (BFGS), the solver converged to local minima. The assumption is that this is owing to the optimal control being close to being ill-posed in its current form, but further investigation into this is beyond the scope of the investigation in this paper.

Deep learning often uses the \Ac*{relu} activation function, defined by $\text{ReLU}(x) = \max(0, x)$. However, \Ac*{relu} is not a smooth activation function, and the derivative is zero for $x<0$. This makes derivative-based optimization more difficult. Instead, we use the Leaky \Ac*{relu} activation function, 

$\text{LeakyReLU}(x) = 
\begin{cases} 
x & \text{if } x > 0, \\
\alpha x & \text{otherwise}.
\end{cases}$, \\ which is also not smooth, but the derivative is not zero for $x<0$. We also use the Mish activation function \cite{misra2019mish}, $$\text{Mish}(x) = x \tanh{(\ln{(1+e^x)})},$$ which is both smooth and the derivative is defined $\forall x \in \mathbb{R}$.


\section{Methodology}\label{sec:methodology}


Our methodology can be summarized as follows:
\begin{itemize}
    \item We train \Ac*{drl} policies using the Stable-Baselines3 \cite{stable-baselines3} implementation of the \ac*{sac} algorithm \cite{haarnoja2018soft} and collect data sets using a modified version of the LunarLander-v2 environment from OpenAI Gymnasium \cite{towers_gymnasium_2023}. We evaluate the policies over 1000 episodes and select policies with a high average but low minimum reward. The idea is that this will result in policies that generally perform well but fail in certain situations. 
    \item From the generated data sets, we identify trajectories (episodes) where the policy crashes the vessel or fails to land in time.
    \item We use \Ac*{pacmap} to identify successful landing episodes that are close in latent space to the unsuccessful ones.
    \item We then use optimization to plan a sequence of actions to move the state of the unsuccessful episode to a state close to the goal location in latent space derived from a successful episode.
    \item Finally, we apply the planned sequence of actions to the environment, starting from the unsuccessful episode, and observe the outcomes.
\end{itemize}

\begin{figure}
    \centering
    \vspace{0.2cm}
    \includegraphics[width=0.95\linewidth]{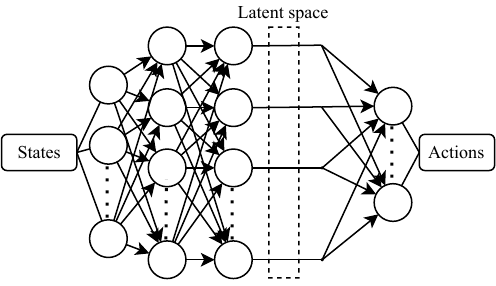}
    \caption{Policy latent space illustration.}
    \label{fig:latent_space_network}
\end{figure}

\subsection{Problem formulation}

The LunarLander-v2 environment is a reinforcement learning environment that emulates landing a spacecraft on the moon. We use the continuous version of this environment for our experiments. The agent controls the spacecraft's main engine and side thrusters using a two-dimensional continuous action. The agent receives a reward for each step based on the lander's position, speed, orientation, and engine usage: closer proximity to the landing pad and slower movement increase rewards while tilting away from horizontal and using engines decreases them. Landing leg contact adds points, whereas firing side or main engines deduct points per frame. The agent receives an additional reward of -100 for crashing or +100 for landing safely. An episode is considered solved if it scores a minimum of 200 points. 

Using the LunarLander-v2 environment for the experiments, we want to solve the following optimization problem:

\begin{equation}\label{eq:abstract_optimal_control_statement}
\begin{aligned}
\min_{\mathbf{x}, \mathbf{u}} \quad & ||B_{D}-\pi_L(x_T)||\\
\textrm{s.t.} \quad & x_{t+1} = f_e(x_t, u_t)\\
  & h(x_t, u_t, t) \leq 0,    \\
\end{aligned}
\end{equation}
where $B_D$ is the set of latent space points in a specific latent space hypervolume, hypothesized to correspond to a desired behavioral mode, $ \pi_L$ is a function mapping states to latent space, i.e., all layers of the policy network except the last layer, as illustrated in \Cref{fig:latent_space_network}. $h(x_t, u_t, t)$ describes the path constraints, i.e., state and action limits in the \ac*{rl} environment. $f_e = \dot{\mathbf{x}}$ is an estimation of the dynamical model of the LunarLander-v2:

\begin{equation*}
f_e = \left[ \begin{aligned}
& x_3 \\
& x_4 \\
& \frac{-\sin(x_5) \cdot u_1 \cdot f_m + \cos(x_5) \cdot u_2 \cdot f_s}{m} \\
& \frac{\cos(x_5) \cdot u_1 \cdot f_m + \sin(x_5) \cdot u_2 \cdot f_s - g}{m} \\
& x_6 \\
& c_{torque} \frac{u_2}{I}
\end{aligned} \right],
\end{equation*}

\vspace{2pt}
where $x_1$ and $x_2$ are the x and y positions, $x_3$ and $x_4$ are the x and y linear velocities, $x_5$ is the angle, and $x_6$ is the angular velocity. $I$ is the inertia, $m$ is the mass of the lander, $g$ is the gravitational constant, and $c_{torque}$ is a constant transforming the force to torque.

The optimal control formulation in (\ref*{eq:abstract_optimal_control_statement}) implies that we want to minimize the distance between the latent space representation of the environment state and $B_D$, the latent space hypervolume containing the desired behavioral mode. However, it is challenging to define $B_D$. Therefore, we instead optimize towards a specific point, which we hypothesize to be within $B_D$. We choose this latent space point using \Ac*{pacmap}, and at the same time, choose $x_0$, the state we initialize the optimization from. 

To solve the optimization problem, we use IPOPT \cite{wachter2006implementation} in CasADI \cite{Andersson2018}. To use PyTorch \cite{paszke2019pytorch} models in CasADI, we use the L4CasADi Python library \cite{salzmann2023l4casadi}. The objective function is the L2 norm of the difference between the desired latent space point and the predicted latent space location at the end of the optimization horizon, as shown in (\ref*{eq:abstract_optimal_control_statement}).

\subsection{Simplifying environments}

To simplify the optimal control task, we took some steps to modify the LunarLander-v2 environment. In the original LunarLander environment, noise is added to the main thruster meant to emulate the thruster's dispersion. We removed this noise, making the actions' influence on the environment deterministic. We also removed the random force applied to the lander in the beginning of the episodes. Instead, we implemented a procedure for initializing the lander in a specific state for a more straightforward recreation of the particular situations we later want to investigate. We also allowed the lander to fire its side thrusters with less than 50\% thrust, making the action space continuous.

\section{Results and Discussion}\label{sec:results_discussion}

This section shows results from using optimal control to switch the behavioral mode of a \Ac*{drl} agent controlling the lunar lander task described above. We use two different \Ac*{drl} policies, summarized in \Cref{tab:policies}. As detailed in the table, both policies are evaluated using the environment and methodology described in \Cref{sec:methodology} over 1000 episodes.
\begin{table}
\centering
\vspace{0.5cm}

\begin{tabular}{lll}
\hline
                                                                               & Policy A                                                                  & Policy B                                                                   \\ \hline
Activation func                                                                & \begin{tabular}[c]{@{}l@{}}torch.nn.\\ Mish\end{tabular}                  & \begin{tabular}[c]{@{}l@{}}torch.nn.\\ LeakyReLU\end{tabular}              \\
\begin{tabular}[c]{@{}l@{}}Avg. reward over \\ 1000 eval episodes\end{tabular} & 134.922                                                                   & 206                                                                        \\
                                                                               &                                                                           &                                                                            \\
\begin{tabular}[c]{@{}l@{}}Hidden layer \\ architecture\end{tabular}           & \begin{tabular}[c]{@{}l@{}}Two dense layers \\ with 64 nodes\end{tabular} & \begin{tabular}[c]{@{}l@{}}Two dense layers \\ with 128 nodes\end{tabular}
\end{tabular}
\caption{Data describing policies.}
\label{tab:policies}
\end{table}

\subsection{Case Study 1}

\def\lefttrim{1}
\def\lowertrim{3.5}
\def\righttrim{4}
\def\uppertrim{0}

\begin{figure}
    \centering
    \includegraphics{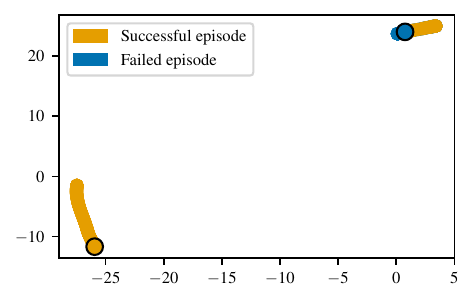}
    \caption{Case Study 1: \Ac*{pacmap} low-dimensional embedding showing latent space initial and goal location.}
    \label{fig:mish_pacmap}
\end{figure}

\begin{figure}
    \centering
    \vspace{0.25cm}
    \begin{subfigure}[b]{0.23\textwidth}
        \centering
        \includegraphics[width=\textwidth,trim={\lefttrim cm \lowertrim cm \righttrim cm \uppertrim cm},clip]{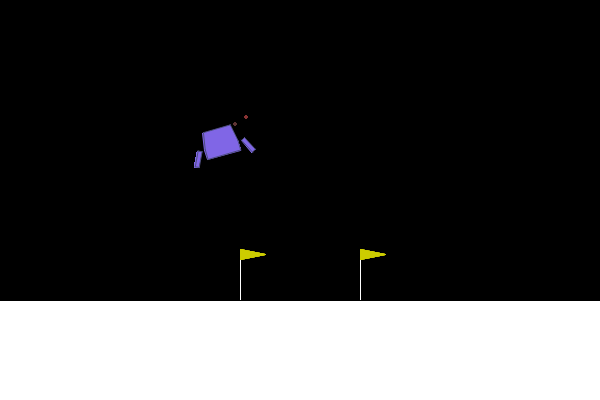}
        \caption{Initial state in situation}
        \label{fig:situation1_image_init}
    \end{subfigure}
    \hfill
    \begin{subfigure}[b]{0.23\textwidth}
        \centering
        \includegraphics[width=\textwidth,trim={\lefttrim cm \lowertrim cm \righttrim cm \uppertrim cm},clip]{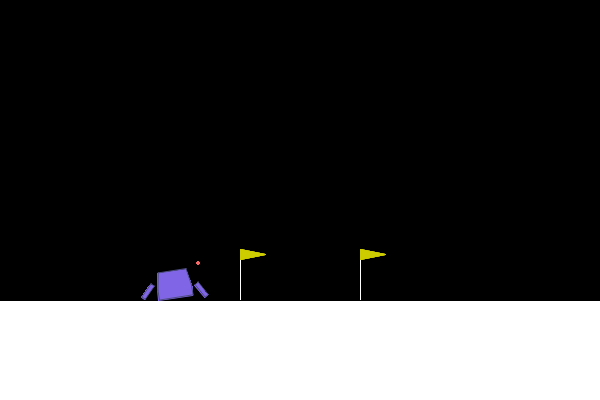}
        \caption{Failure by policy}
        \label{fig:situation1_image_failure}
    \end{subfigure}
    \vskip\baselineskip 
    \begin{subfigure}[b]{0.23\textwidth}
        \centering
        \includegraphics[width=\textwidth,trim={\lefttrim cm \lowertrim cm \righttrim cm \uppertrim cm},clip]{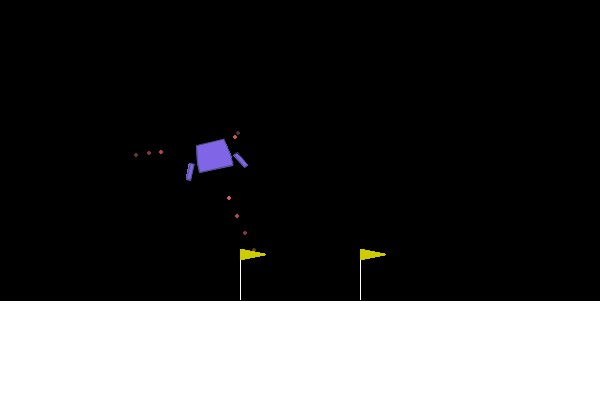}
        \caption{State reached by optimization}
        \label{fig:situation1_image_switched}
    \end{subfigure}
    \hfill
    \begin{subfigure}[b]{0.23\textwidth}
        \centering
        \includegraphics[width=\textwidth,trim={\lefttrim cm \lowertrim cm \righttrim cm \uppertrim cm},clip]{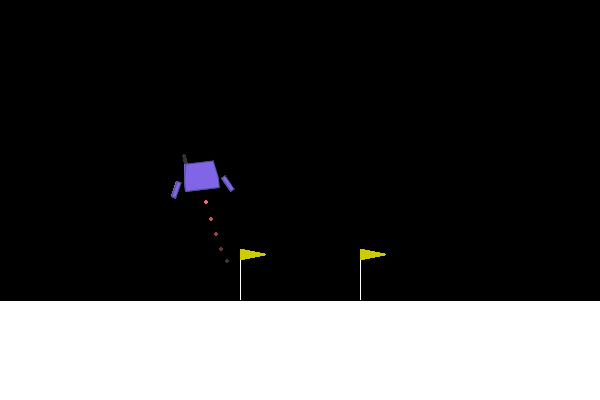}
        \caption{Goal latent space point's state}
        \label{fig:situation1_image_desired}
    \end{subfigure}
    \caption{Case Study 1.}
    \label{fig:situation1_images}
\end{figure}

\begin{figure}
    \centering
    \includegraphics[width=.87\linewidth]{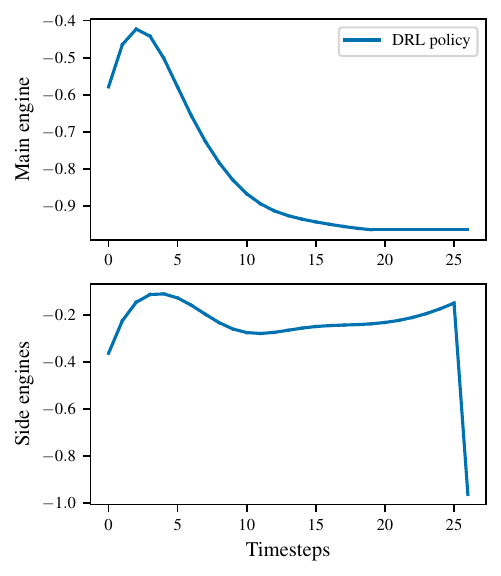}
    \caption{Case Study 1: Actions when chosen by the policy during the failed episode.}
    \label{fig:situation1_orig_actions}
\end{figure}

\begin{figure}
    \centering
    \includegraphics[width=.87\linewidth]{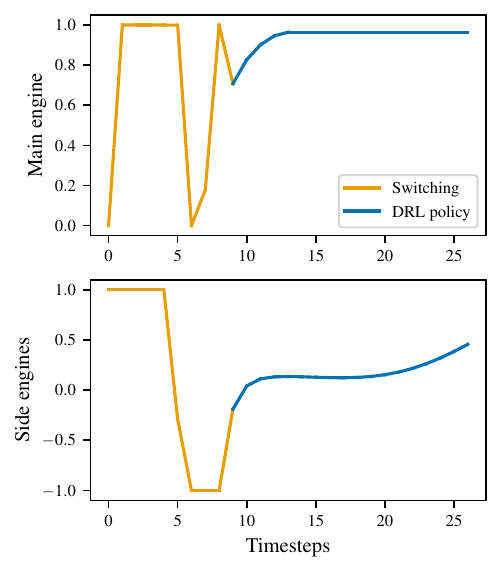}
    \caption{Case Study 1: Actions when switching behavior.}
    \label{fig:situation1_switch_actions}
\end{figure}

\begin{figure}
    \centering
    \includegraphics[width=.87\linewidth]{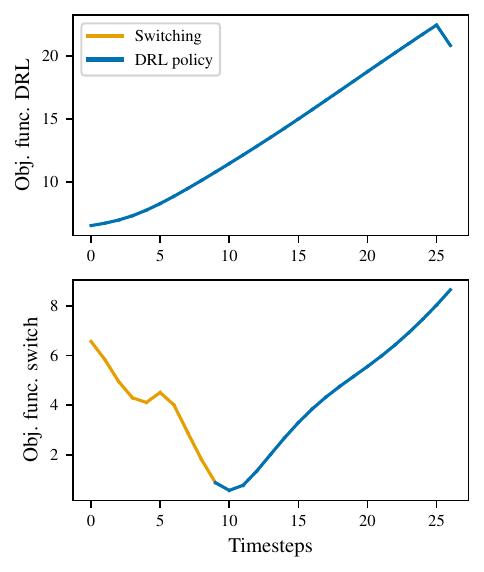}
    \caption{Case Study 1: Objective function difference.}
    \label{fig:situation1_obj_func}
\end{figure}

During this first situation, Policy A controls the vessel. When the policy controls the vessel, the vessel falls right down and hits the ground, as seen in \Cref{fig:situation1_image_failure}. From \Cref{fig:situation1_orig_actions}, we can see that the main engine action is negative for the entirety of the situation when Policy A controls the vessel. This means that the main engine is never firing. We hypothesize that this behavior occurs because the policy never encountered a similar situation during training and, therefore, has a subpar behavior in this region of the latent space. We find another episode very close in latent space, where the policy successfully lands the agent. The main difference between these two episodes is that the vessel's angle is closer to the neutral position during the successful episode. We visualize the latent space points for these two episodes in \Cref{fig:mish_pacmap} using \Ac*{pacmap}. The \Ac*{pacmap} algorithm clusters both the successful episode shown in \mbox{orange \fcolorbox{black}{orange}{\rule{0pt}{6pt}\rule{6pt}{0pt}} } and the unsuccessful episode shown in blue \fcolorbox{black}{blue}{\rule{0pt}{6pt}\rule{6pt}{0pt}} in the top-right corner at the start, so it is difficult to distinguish their start using this embedding. When the vessel starts to fire the main engine and correctly lands the vessel in the successful episode, the latent space representation jumps to the bottom left corner of the embedding. This indicates that the policy changes its behavioral mode here. We intervene in the failed episode at the latent space location illustrated in \Cref{fig:mish_pacmap} by the blue circle with a black border. This latent space location corresponds to the initial situation, as seen in \Cref{fig:situation1_image_init}. We want to switch the policy's behavioral mode during the failed episode to the behavioral mode visualized in the bottom left corner in \Cref{fig:mish_pacmap}. We use the latent space location visualized by the orange circle with a black border, corresponding to \Cref{fig:situation1_image_desired}, as our goal latent space location and use the optimization procedure detailed in \Cref{sec:methodology} to find a sequence of actions which pushes the latent space representation of the state towards that location. The state reached can be seen in \Cref{fig:situation1_image_switched}. 

When comparing the actions chosen by the policy during the failed episode, seen in \Cref{fig:situation1_orig_actions} with the actions selected by the optimization procedure when switching the behavior, seen in \Cref{fig:situation1_image_switched}, we can see that the way to switch the behavior involves first firing the main engine, deaccelerating the vessel. When comparing the side engine action, we can also notice that the optimization procedure fires the side engine at max. This would lead to the vessel's angle decreasing, indicating that the hypothesis regarding the vessel's angle being important for which of these behavioral modes the policy was correct.

\Cref{fig:situation1_obj_func} shows how the optimization procedure pushes the environment towards a state much closer to the desired latent space location than ever reached during the failed episode.

The cumulative reward throughout the episode changes from -115 in the failed episode to +241 in the episode where we switch the behavioral mode using optimal control.

\subsection{Case Study 2}

\begin{figure}
    \centering
    \vspace{0.25cm}
    \begin{subfigure}[b]{0.23\textwidth}
        \centering
        \includegraphics[width=\textwidth,trim={\lefttrim cm \lowertrim cm \righttrim cm \uppertrim cm},clip]{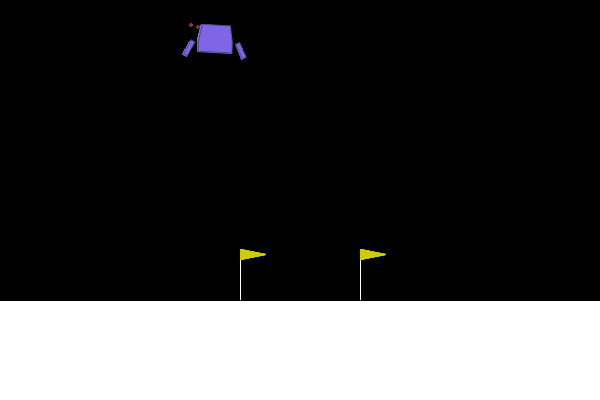}
        \caption{Initial state in situation}
        \label{fig:situation2_image_init}
    \end{subfigure}
    \hfill
    \begin{subfigure}[b]{0.23\textwidth}
        \centering
        \includegraphics[width=\textwidth,trim={\lefttrim cm \lowertrim cm \righttrim cm \uppertrim cm},clip]{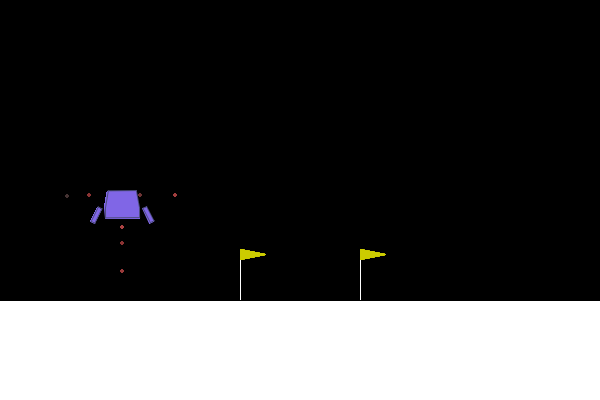}
        \caption{Failure by policy}
        \label{fig:situation2_image_failure}
    \end{subfigure}
    \vskip\baselineskip 
    \begin{subfigure}[b]{0.23\textwidth}
        \centering
        \includegraphics[width=\textwidth,trim={\lefttrim cm \lowertrim cm \righttrim cm \uppertrim cm},clip]{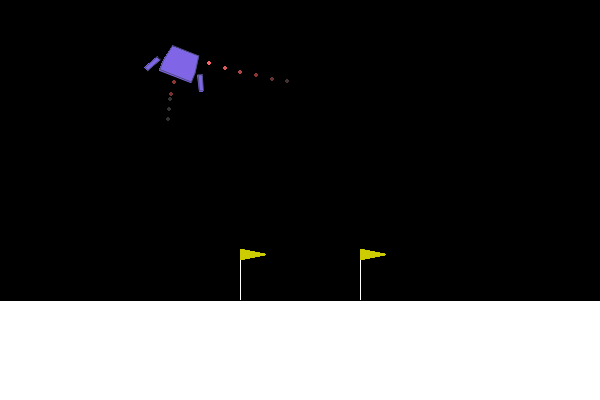}
        \caption{State reached by optimization}
        \label{fig:situation2_image_switched}
    \end{subfigure}
    \hfill
    \begin{subfigure}[b]{0.23\textwidth}
        \centering
        \includegraphics[width=\textwidth,trim={\lefttrim cm \lowertrim cm \righttrim cm \uppertrim cm},clip]{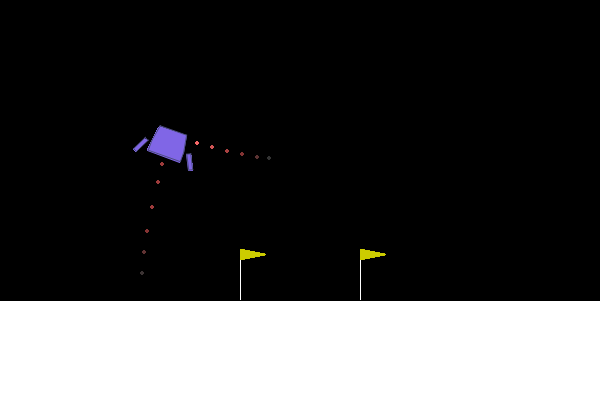}
        \caption{Goal latent space point's state}
        \label{fig:situation2_image_desired}
    \end{subfigure}
    \caption{Case Study 2.}
    \label{fig:situation2_images}
\end{figure}

\begin{figure}
    \centering
    \includegraphics[width=.87\linewidth]{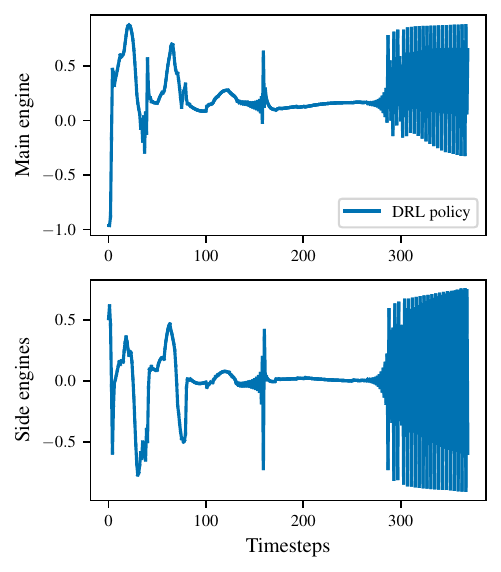}
    \caption{Case Study 2: Actions when chosen by policy.}
    \label{fig:situation2_orig_actions}
\end{figure}

\begin{figure}
    \centering
    \includegraphics[width=.87\linewidth]{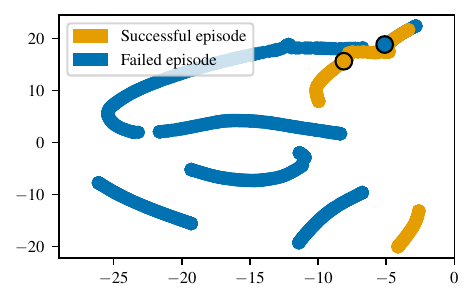}
    \caption{Case Study 2: \Ac*{pacmap} low-dimensional embedding showing latent space initial and goal location.}
    \label{fig:leaky_pacmap}
\end{figure}

\begin{figure}
    \centering
    \includegraphics[width=.87\linewidth]{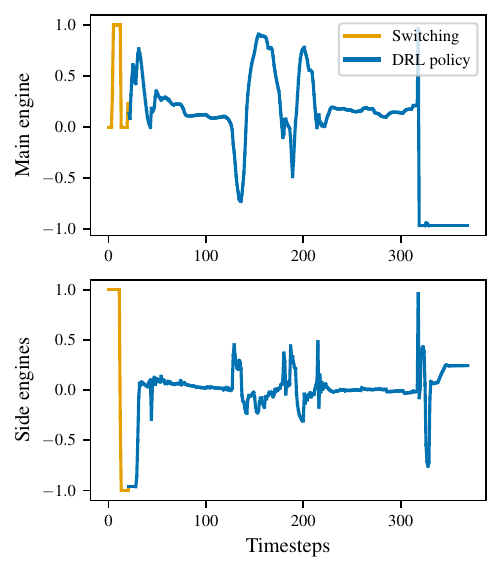}
    \caption{Case Study 2: Actions when switching behavior.}
    \label{fig:situation2_switch_actions}
\end{figure}

\begin{figure}
    \centering
    \includegraphics[width=.87\linewidth]{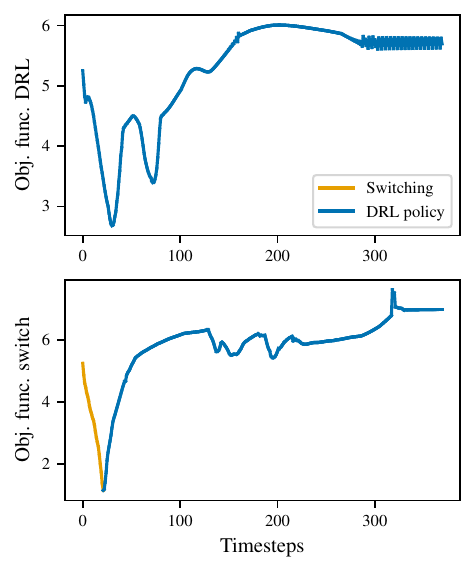}
    \caption{Case Study 2: Objective function difference.}
    \label{fig:situation2_obj_func}
\end{figure}

In the second situation, Policy B controls the vessel. During this situation, the policy gets stuck switching cycling between four different behavioral modes, with the result being that the vessel never lands, and just hovers in place, seen in \Cref{fig:situation2_image_failure}, and indicated by the plot of the actions in \Cref{fig:situation2_orig_actions}. These four behavioral modes can be seen in the \Ac*{pacmap} two-dimensional embedding of the policy's latent space in \Cref{fig:leaky_pacmap} as the four separate blue regions of points starting in the middle of the plot and going down. 

We find an episode with an initial similarity in its latent space representation, shown in orange in \Cref{fig:leaky_pacmap}. However, in this episode, the policy does not get stuck in a loop but lands the vessel successfully. Similar to in Case Study 1, we pick the initialize the optimization at the latent space location corresponding to the blue circle with black border in \Cref{fig:leaky_pacmap}, corresponding to \Cref{fig:situation2_image_init}, and plan a sequence of actions to move the latent space representation of the state towards the orange circle with black border, corresponding to \Cref{fig:situation2_image_desired}. 

\Cref{fig:situation2_obj_func} shows how applying the actions from the optimization, seen in \Cref{fig:situation2_switch_actions}, moves the latent space representation of the environment's state towards a location that is closer to the desired latent space location than what letting Policy B run from the initial state. The state space version of this latent space location can be seen in \Cref{fig:situation2_image_switched}.

The cumulative reward throughout the episode changes from -58 in the failed episode to +228 in the episode where we switch the behavioral mode using optimal control.


\subsection{Moving successful behavioral mode to failed behavioral mode}

To further demonstrate the ability to change the behavioral mode of the policy, we do two experiments where we do the opposite of what we did in Case Study 1 and Case Study 2. That is, we move the policy from a successful behavioral mode to an unsuccessful behavioral mode.

For Case Study 1, we initialize in the successful episode, where the vessel has a small enough angle that the policy eventually moves to the behavioral mode in the bottom left of \Cref{fig:mish_pacmap}. We intervene and use the optimization procedure to push the latent space representation of the state towards the failed episode, with the result being that the vessel crashes, decreasing the cumulative reward throughout the episode from +245 to -116. 

For Case Study 2, we initialize the agent in the behavioral mode, which does not eventually reach where the policy gets stuck, switching between four different behavioral modes. However, we intervene and use the optimization procedure to push the latent space representation towards that region of the latent space, with the result being that the policy gets stuck similarly to in the failed episode, decreasing the cumulative reward throughout the episode from +233 to -79.


\section{Conclusion}\label{sec:conclusion}

This paper has introduced a practical method for manipulating the behavioral modes of \Ac*{drl} policies using optimal control in the policies' latent space. Our approach demonstrates a direct way to influence the behavior of DRL policies, showing how we can switch a policy's behavioral mode to affect policy performance in simulated environments, as demonstrated with our lunar lander case study.

This work uses an analytical model of the \Ac*{drl} environment. However, an analytical model is not readily available for many more complex \Ac*{drl} environments. Further work could, therefore, involve approximating a model of the environment and then using this approximated model for optimization.




\section*{Acknowledgment}
The Research Council of Norway supported this work through the EXAIGON project, project number 304843.

\bibliographystyle{IEEEtran}
\bibliography{mylib}

\begin{thebibliography}{10}
\providecommand{\url}[1]{#1}
\csname url@samestyle\endcsname
\providecommand{\newblock}{\relax}
\providecommand{\bibinfo}[2]{#2}
\providecommand{\BIBentrySTDinterwordspacing}{\spaceskip=0pt\relax}
\providecommand{\BIBentryALTinterwordstretchfactor}{4}
\providecommand{\BIBentryALTinterwordspacing}{\spaceskip=\fontdimen2\font plus
\BIBentryALTinterwordstretchfactor\fontdimen3\font minus \fontdimen4\font\relax}
\providecommand{\BIBforeignlanguage}[2]{{%
\expandafter\ifx\csname l@#1\endcsname\relax
\typeout{** WARNING: IEEEtran.bst: No hyphenation pattern has been}%
\typeout{** loaded for the language `#1'. Using the pattern for}%
\typeout{** the default language instead.}%
\else
\language=\csname l@#1\endcsname
\fi
#2}}
\providecommand{\BIBdecl}{\relax}
\BIBdecl

\bibitem{badia2020agent57}
A.~P. Badia, B.~Piot, S.~Kapturowski, P.~Sprechmann, A.~Vitvitskyi, Z.~D. Guo, and C.~Blundell, ``Agent57: Outperforming the atari human benchmark,'' in \emph{International conference on machine learning}.\hskip 1em plus 0.5em minus 0.4em\relax PMLR, 2020, pp. 507--517.

\bibitem{kiran2021deep}
B.~R. Kiran, I.~Sobh, V.~Talpaert, P.~Mannion, A.~A. Al~Sallab, S.~Yogamani, and P.~P{\'e}rez, ``Deep reinforcement learning for autonomous driving: A survey,'' \emph{IEEE Transactions on Intelligent Transportation Systems}, vol.~23, no.~6, pp. 4909--4926, 2021.

\bibitem{salzmann2023real}
T.~Salzmann, E.~Kaufmann, J.~Arrizabalaga, M.~Pavone, D.~Scaramuzza, and M.~Ryll, ``Real-time neural mpc: Deep learning model predictive control for quadrotors and agile robotic platforms,'' \emph{IEEE Robotics and Automation Letters}, vol.~8, no.~4, pp. 2397--2404, 2023.

\bibitem{wang2021understanding}
Y.~Wang, H.~Huang, C.~Rudin, and Y.~Shaposhnik, ``{Understanding How Dimension Reduction Tools Work: An Empirical Approach to Deciphering t-SNE, UMAP, TriMap, and PaCMAP for Data Visualization},'' \emph{J. Mach. Learn. Res.}, vol.~22, no. 201, pp. 1--73, 2021.

\bibitem{wachter2006implementation}
A.~W{\"a}chter and L.~T. Biegler, ``On the implementation of an interior-point filter line-search algorithm for large-scale nonlinear programming,'' \emph{Mathematical programming}, vol. 106, pp. 25--57, 2006.

\bibitem{misra2019mish}
D.~Misra, ``Mish: A self regularized non-monotonic activation function,'' \emph{arXiv preprint arXiv:1908.08681}, 2019.

\bibitem{stable-baselines3}
A.~Raffin, A.~Hill, A.~Gleave, A.~Kanervisto, M.~Ernestus, and N.~Dormann, ``Stable-baselines3: Reliable reinforcement learning implementations,'' \emph{J. Mach. Learn. Res.}, vol.~22, no. 268, pp. 1--8, 2021.

\bibitem{haarnoja2018soft}
T.~Haarnoja, A.~Zhou, K.~Hartikainen, G.~Tucker, S.~Ha, J.~Tan, V.~Kumar, H.~Zhu, A.~Gupta, P.~Abbeel \emph{et~al.}, ``Soft actor-critic algorithms and applications,'' \emph{arXiv preprint arXiv:1812.05905}, 2018.

\bibitem{towers_gymnasium_2023}
\BIBentryALTinterwordspacing
M.~Towers, J.~K. Terry, A.~Kwiatkowski, J.~U. Balis, G.~d. Cola, T.~Deleu, M.~Goulão, A.~Kallinteris, A.~KG, M.~Krimmel, R.~Perez-Vicente, A.~Pierré, S.~Schulhoff, J.~J. Tai, A.~T.~J. Shen, and O.~G. Younis, ``Gymnasium,'' Mar. 2023. [Online]. Available: \url{https://zenodo.org/record/8127025}
\BIBentrySTDinterwordspacing

\bibitem{Andersson2018}
J.~A.~E. Andersson, J.~Gillis, G.~Horn, J.~B. Rawlings, and M.~Diehl, ``{CasADi} -- {A} software framework for nonlinear optimization and optimal control,'' \emph{Mathematical Programming Computation}, 2018.

\bibitem{paszke2019pytorch}
A.~Paszke, S.~Gross, F.~Massa, A.~Lerer, J.~Bradbury, G.~Chanan, T.~Killeen, Z.~Lin, N.~Gimelshein, L.~Antiga \emph{et~al.}, ``Pytorch: An imperative style, high-performance deep learning library,'' \emph{Advances in neural information processing systems}, vol.~32, 2019.

\bibitem{salzmann2023l4casadi}
T.~Salzmann, J.~Arrizabalaga, J.~Andersson, M.~Pavone, and M.~Ryll, ``{Learning for CasADi: Data-driven Models in Numerical Optimization},'' 2023.

\end{thebibliography}

\end{document}